\documentclass[a4paper]{article}
\usepackage{hyperref}
\usepackage[labelsep=period]{caption}
\usepackage{array, bigdelim, makecell}  
\usepackage{comment}
\usepackage{array}
\usepackage{siunitx}
\usepackage{xcolor}
\usepackage{enumitem}
\usepackage{graphicx}
\usepackage{multirow}
\usepackage{rotating}
\usepackage{amsmath}
\usepackage{cleveref}

\hypersetup{colorlinks = true, allcolors = blue}

\usepackage{graphicx}
\usepackage{multicol}
\usepackage{algcompatible}
\usepackage[ruled,vlined]{algorithm2e}
\usepackage{listings}
\usepackage{float}
\usepackage[noend]{algpseudocode}
\usepackage{caption}
\usepackage[labelformat=simple]{subcaption}

\usepackage{multicol}
\usepackage[noend]{algpseudocode}
\usepackage[section]{placeins}

\usepackage{array}
\newcolumntype{P}[1]{>{\centering\arraybackslash}p{#1}}
\newcolumntype{L}{>{\centering\arraybackslash}m{6cm}}\large\LARGE\Huge\normalsize
\newcolumntype{C}{>{\centering\arraybackslash}m{3cm}}

\usepackage[margin=1in]{geometry}
\usepackage[utf8]{inputenc}

\usepackage[sort&compress,numbers,square]{natbib}
\title{Biomarker Gene Identification for Breast Cancer Classification}
\author{Sheetal Rajpal$^1$ \and Ankit Rajpal$^1$ \thanks{Correponding Author} \and Manoj Agarwal$^2$ \and Naveen Kumar $^1$}

\date{
	$^1$Department of Computer Science, University of Delhi \\ \texttt{sheetal.rajpal.09@gmail.com, arajpal@cs.du.ac.in, nkumar@cs.du.ac.in}\\%
	$^2$Department of Computer Science, Hans Raj College, University of Delhi \\ \texttt{manoj.agarwal@hrc.du.ac.in}\\
}

\begin{document}
\maketitle

\begin{abstract}
\text{}\\
\textbf{BACKGROUND:} Breast cancer has emerged as one of the most prevalent cancers among women leading to a high mortality rate. Due to the heterogeneous nature of breast cancer, there is a need to identify differentially expressed genes associated with breast cancer subtypes for its timely diagnosis and treatment.\\
\textbf{OBJECTIVE:} To identify a small gene set for each of the four breast cancer subtypes that could act as its signature, the paper proposes a novel algorithm for gene signature identification.\\
\textbf{METHODS:} The present work uses interpretable AI methods to investigate the predictions made by the deep neural network employed for subtype classification to identify biomarkers using the TCGA breast cancer RNA Sequence data.\\
\textbf{RESULTS:} The proposed algorithm led to the discovery of a set of 43 differentially expressed gene signatures. We achieved a competitive average 10-fold accuracy of 0.91, using neural network classifier. Further, gene set analysis revealed several relevant pathways, such as GRB7 events in ERBB2 and p53 signaling pathway. Using the Pearson correlation matrix, we noted that the subtype-specific genes are correlated within each subtype.\\
\textbf{CONCLUSIONS:} The proposed technique enables us to find a concise and clinically relevant gene signature set.
\end{abstract}

\section{Introduction}\label{sec:intro}

\par Cancer is a complex heterogeneous disorder caused by uncontrolled growth of abnormal cells that may affect several parts of the body. Amongst women, breast cancer is still the major cause of mortality causing millions of deaths every year and observed the highest incidence rate in 2020 \cite{sung2021global}. The primary cause of the disease is the variations at a molecular and cellular level with only 5\% of the cases being hereditary \cite{callahan1992somatic}. The advancement of next-generation sequencing techniques has offered us the opportunity to analyze the disease at the transcriptome level, in form of gene expression data, which was not possible earlier \cite{reis2009next,sotiriou2009gene}. Although gene expression data analysis provides potential advantages, however, handling high dimensional and imbalanced small-size datasets is a challenging task.

\par Breast cancer being a heterogeneous disease, it is pertinent to classify it into several subtypes. In this regard, several classification categories such as histological grading, TNM staging, and molecular subtyping have been proposed in literature\cite{parker2009supervised}. Amongst all categories, molecular subtyping is more promising towards clinical and prognostic outcomes. It defines four main subtypes, mainly, Basal, Her2, Luminal A, and Luminal B. Though these subtypes have been established through IHC defined markers (Estrogen Receptor (ER), Progesterone Receptor (PR), and human epidermal growth factor receptor 2 (Her2)) as well as PAM50, the latter is considered to be the gold standard because of being more significantly linked with outcomes \cite{parker2009supervised,cheang2009ki67,perou2000molecular,sorlie2003repeated,kim2019discordance,eccles2013critical}.

\par Selecting an effective breast cancer therapy method requires finding the proper subtype and identifying differentially expressed genes associated with breast cancer subtypes \cite{prat2015clinical,dai2015breast}. Thus, different statistical methods have been proposed for biomarker discovery. For evaluating their classification performance, several machine learning techniques are widely used in literature \cite{wu2017pathways,gao2019deepcc,list2014classification}. However, with the advances in machine learning techniques, cutting-edge deep learning algorithms appear to be more promising, outperforming the outputs provided by traditional machine learning methodologies. \cite{zhang2017machine,korotcov2017comparison}. In this manuscript, we aim to propose an algorithm for the discovery of gene signature, by analyzing a deep neural network, that may potentially serve as a set of target genes for devising a therapy.

\par In literature, several attempts for discovering gene signature set capable of differentiating molecular subtypes have focused on using IHC marker defined subtypes \cite{wu2017pathways,graudenzi2017pathway,sherafatian2018tree,tao2019classifying}. Because of the superiority of PAM50 over IHC defined subtypes as discussed before, several research groups have considered PAM50 subtypes for classification and biomarker genes identification \cite{list2014classification,zhang2017novel,gao2019deepcc}. In this direction, we could find only one comparable work of Zhang et al. related with that of ours with respect to the four-class classification problem. Zhang et al. \cite{zhang2018lncrna} investigated the contribution of long non-coding RNAs (lncRNAs) from the TCGA RNAseq data and identified a set of lncRNAs features that were used in conjunction with selected coding genes and PAM50 genes for breast cancer subtype prediction and obtained prediction accuracy of 0.956 using 106 genes. However, since the higher accuracy is achieved at the expense of a high number of genes, authors have obtained reduced representation of 36 genes (coding +non-coding) and the 50 gene set signature achieving 0.876 and 0.885 for the four-class classification problem. In this study, the Recursive 1-Norm SVM was used for feature selection and 2-Norm SVM for classification. 

\par We have proposed an algorithm for gene signature identification for each of the four subtypes of breast cancer \cite{rajpal2021triphasic}. The algorithm investigates the deep neural network classifier built for subtype classification using backpropagation methods. Using the TCGA breast cancer RNA Sequence data, the proposed algorithm led to the discovery of 43 differentially expressed gene signature set. Using these 43 genes as the input to a neural network classifier, we obtained mean 10-fold test accuracy of 0.91.

\par In comparison to the literature, our framework is a pure deep learning approach in which the classifier neural network developed is leveraged for further analysis through Innvestigate tool for determining potential biomarker genes for differentiating between several breast cancer subtypes. Though related studies in the literature have reported different sets of genes, however, our approach demonstrates competitive performance while proposing a concise signature set that appears to be biologically and clinically relevant.

\par The remainder of the paper is arranged as follows: in the second section, we present the gene signature identification algorithm. The next section mentions the experimental details, results, and gene set analysis. Finally, the paper is concluded in \cref{sec:conclusion}. 

\section{Dataset and Methodology} \label{sec2:dataset:method}

\par The section gives the dataset description and presents the Gene Signature 	Identification Algorithm for gene signature identification for each of the four subtypes of breast cancer.

\subsection{Dataset} \label{dataset}

\par We used the TCGA breast cancer dataset from The Cancer Genome Atlas (TCGA) data source for our experiments. It comprises data corresponding to 1218 patients. The accessible information for each patient comprises gene expression data for 20,530 genes as well as clinical information. We examine 1093 samples with PAM50 subtypes available. As a result, we limit our study to 837 patients with four unique subtypes: Luminal A, Luminal B, Basal, and Her2. Among the 837 patients examined, 142 belong to the Basal subtype, 67 belong to the Her2 subtype, 434 belong to the Luminal A category, and 194 belong to the Luminal B category.

\subsection{Gene Signature Identification Algorithm} \label{algorithm}
\par In this section, we aim to identify differentially expressed gene signature set associated with different breast cancer subtypes. Towards this end, we employ a neural network classifier which comprises an autoencoder for finding compact representation followed by another neural network for classification \cite{DLModel}. The trained model used for the classification task is analyzed using the Innvestigate tool. We have used seven explainable AI techniques of innvestigate tool, namely, Guided Backpropagation, Input-t Gradient, LRP-Z, Gradient, Smooth Grad, Integrated Gradient, and LRP Epsilon. These techniques provide explanations for the behavior of a trained network enabling us to identify gene signature. For each of the four subtypes of breast cancer, we selected top 250 most contributing genes for all patients of that subtype. However, we finally shortlisted only those genes which were relevant for atleast 30\% of the patients. Proceeding in this manner, we combined relevant genes identified using different methods into a set termed, candidateSet. With the intent to find a minimal gene signature set, the top 10 subtype-specific genes are computed for each of the subtypes using rank sum test in one vs all setting. 

\begin{figure}[!htbp]
\centering
\includegraphics[width=1.3in,height=3in]{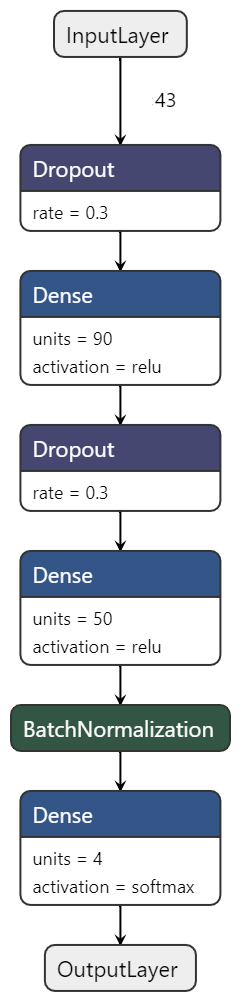}
\caption{Deep Neural Network Architecture for Classifying Breast Cancer Subtypes using 43 Identified Genes}
\label{fig:03c}
\end{figure}

\begin{algorithm}[!htbp]
\caption{\textbf{Gene Signature Identification Algorithm}}

\textbf{Input:} \\
\hspace*{\algorithmicindent} ${\textbf{model}}$: Neural Network model used for the classification task \\
\textbf{Output:} \\
\hspace*{\algorithmicindent} $\textbf{GeneSignature}$: Gene Signature Set \\
	\begin{enumerate}
		\item \begin{math} PAM50Subtypes \leftarrow \{Basal, Her2, Luminal A, Luminal B\} \end{math}
		\item \begin{math} candidateGenes  \leftarrow \{\} \end{math}
    \item for \begin{math} method \end{math} in \{ Explainable AI Methods \} do \text{       }
				\begin{enumerate}
				\item \begin{math}  RelevantGenes[method] =  \{\} \end{math}
				\item for each \begin{math} class  \text{ in } PAM50Subtypes \end{math} do
				\begin{enumerate}
				\item \begin{math} genes[class] \leftarrow top250(class, method, model) \end{math} 
						\item  \begin{math} genes[class] \leftarrow selectFreq(geneSet[class], 0.30 \times length(X[class])) \end{math}
				\item \begin{math} RelevantGenes[method] \leftarrow RelevantGenes[method]  \cup genes[class] \end{math} 
				\end{enumerate} 
				 \item \begin{math} candidateGenes \leftarrow candidateGenes  \cup  RelevantGenes[method] \end{math} 
				\end{enumerate} 
	
	\item \begin{math} GeneSignature \leftarrow \{\} \end{math}

	\item for each $class$ in \begin{math} PAM50Subtypes\end{math} do
			\begin{enumerate}
			\item \begin{math} top10Genes \leftarrow  rankSumTestTop10(candidateGenes, X[class], X[AllSubtypes-subtype]) \end{math}
			\item \begin{math} GeneSignature \leftarrow  GeneSignature \text{  }\cup\text{  } top10Genes \end{math}
			\end{enumerate}

\end{enumerate}
\end{algorithm}

\subsection{Neural Network Classifier Architecture taking discovered gene signature as input} \label{NeuralNet}
\par The candidate set so obtained was evaluated for classification performance using another neural network. This network comprises three dense layers. The first, second, and third layers contain 90, 50, and 4 nodes respectively (\Cref{fig:03c}). A dropout layer is used after the first and second dense layers. While the activation function ReLU is used for the first and second dense layers, the softmax function is used for the third dense layer responsible for classification.

\section{Experimental Details and Results} \label{sec:experimental}
\par This section provides the pre-processing steps applied to the dataset utilized for the experimentation. Subsequent subsections discuss the biomarkers discovered using different explainable AI methods and a comparison of their classification performance with the state-of-the-art techniques. Finally, the clinical relevance of the identified gene signature set is established using the gene set analysis. 

\subsection{Pre-Processing} \label{preprocessing} 
\par Because of the skewed data distribution, we applied z-score normalization to gene expression values for uniform scaling. For experimentation, four PAM50 subtypes namely, Basal, Her2, Luminal A, and Luminal B were provided with numerical values 0, 1, 2, and 3 respectively. To overcome the class imbalance problem, we employed a method known as Synthetic Minority Over-sampling Technique (SMOTE).

\subsection{Gene Signature Selection using Proposed Method and their classification performance} \label{genesetselect}

\par
\begin{figure}[!htbp]
\centering
	\subfloat[ Histogram depicting the count of genes (X-axis) in the signature chosen by different methods of Innvestigate tool (Y-axis) along with their classification performance in terms of accuracy (X-axis) using those genes.]{
	\includegraphics[width=3in,height=2in]{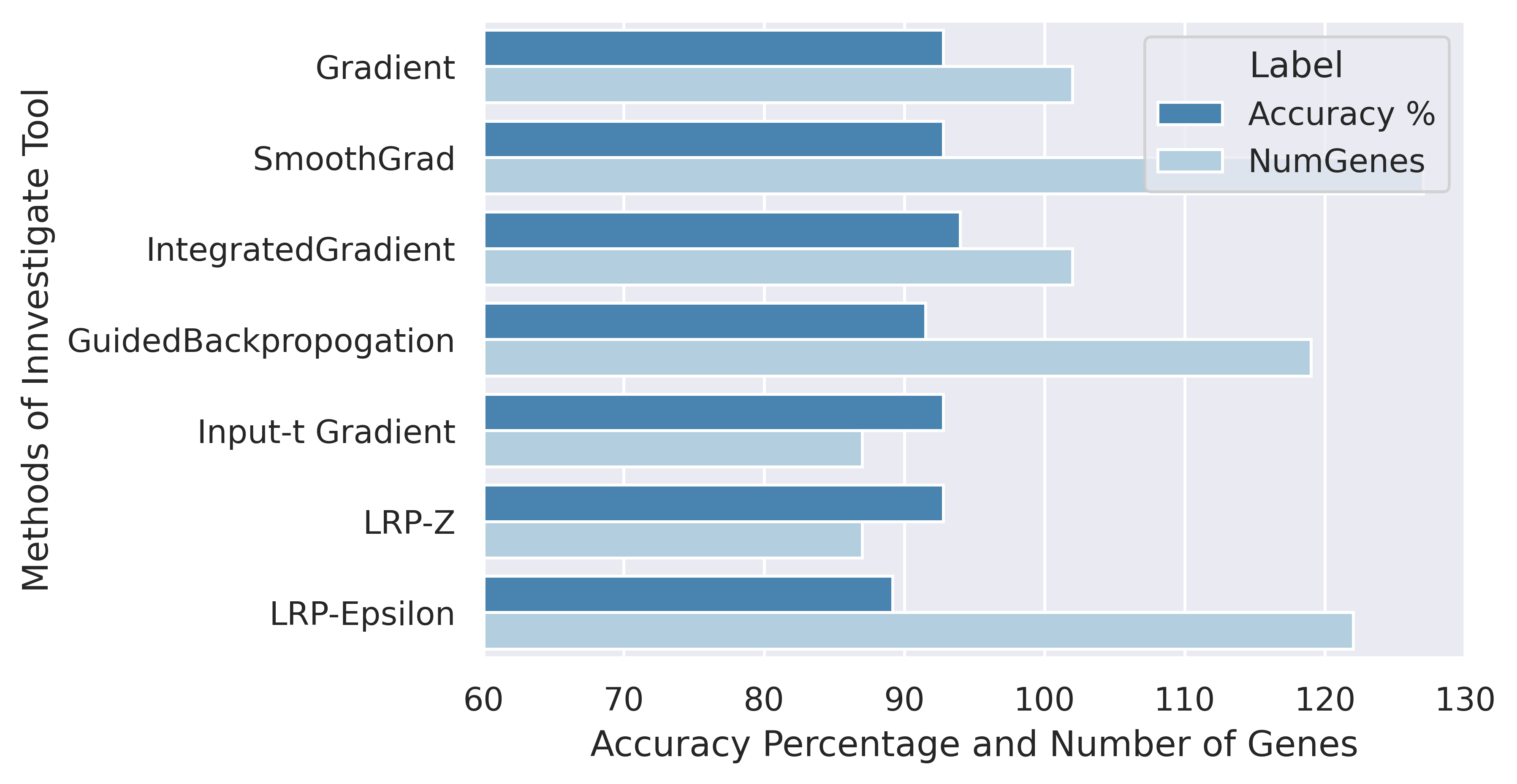}
	\label{fig:05b}}
	\hfill
	\subfloat[The Venn diagram demonstrates the intersection of candidate genes found by the different explainable AI techniques of Innvestigate Tool. The found candidate group shares eight genes.]{
	\includegraphics[width=3in,height=2in]{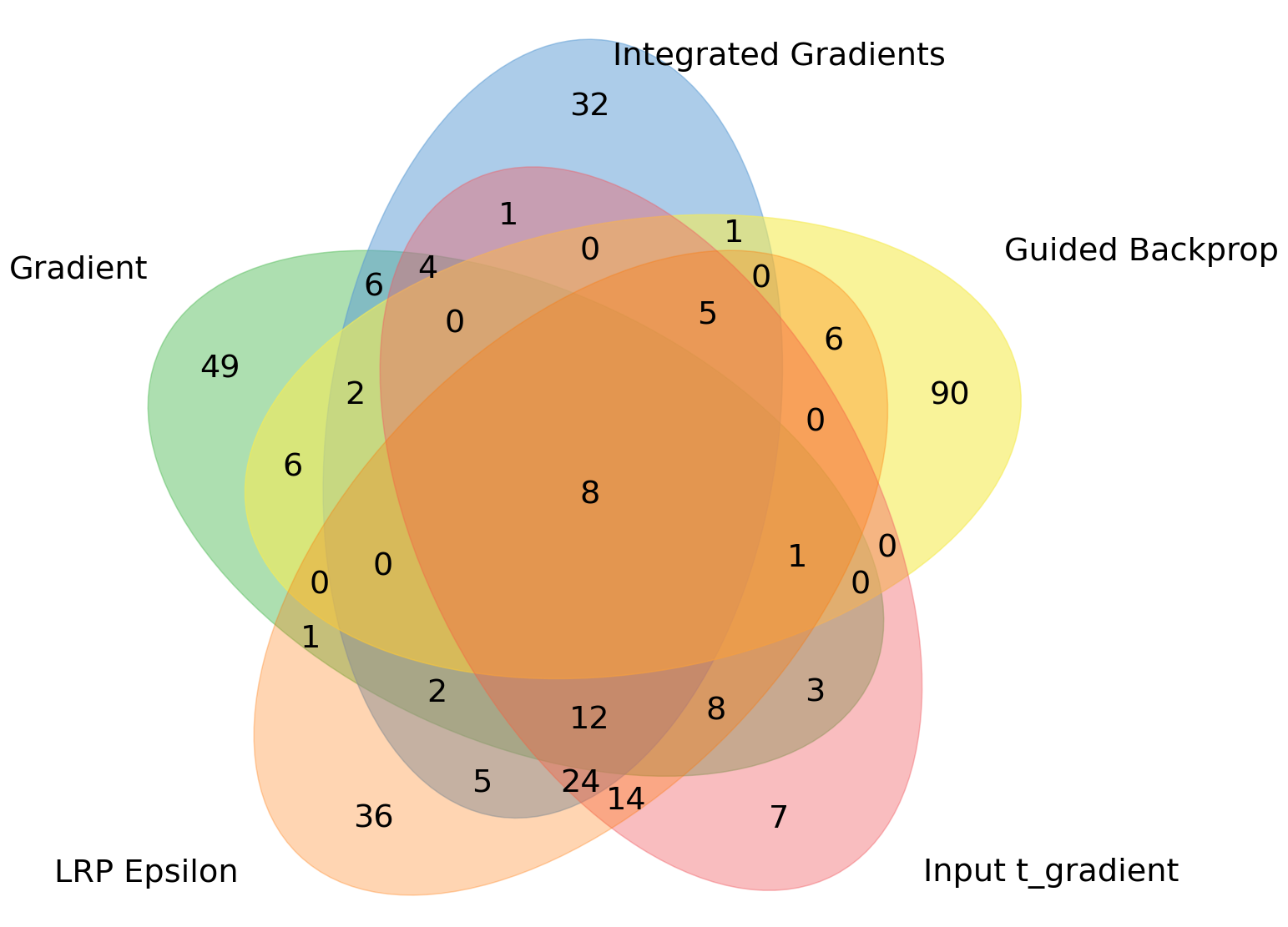}
	\label{fig:05c}}
\caption{ Comparison of results obtained through application of different techniques of Innvestigate tool.}
\label{fig:051}
\end{figure}

 For gene signature identification, we applied the algorithm discussed in \cref{algorithm}. First, the behavior of the trained neural network classifier is analyzed using seven explainable AI methods. A new gene signature is discovered for every individual method. It may be noted from the \Cref{fig:05b} that almost all the techniques identify a list of genes with high classification accuracy ($>$ 0.90). The number of genes identified by different approaches ranges between (87, 127), with the Integrated Gradient method yielding the best accuracy of $0.93$ for a collection of 102 genes.

\par It was noted that Input-t Gradient and LRP-Z identify exactly  the same set of genes. Further, we noted that the set of genes presented by the Gradient method was a subset of genes selected by the SmoothGradient method. As they yield the same accuracy, we only consider the smaller gene subset. So, we selected five methods (please see \Cref{fig:05c}) for comparative study w.r.t gene sets identified by them. The Venn diagram representing the intersection of genes chosen by these five techniques is shown in \Cref{fig:05c}. It should be noted that all of the gene signatures discovered by various approaches share eight genes, notably '\textit{CENPK}', '\textit{TBX10}', '\textit{STARD3}', '\textit{CLCA2}', '\textit{ERBB2}', '\textit{TCAP}', '\textit{GRB7}', and '\textit{MLF1IP}'.

\par We intend to uncover a small set of gene signatures that can distinguish between the four types of breast cancer. Thus, we combined the collection of potential genes discovered by different techniques of the Innvestigate tool, yielding a total of 323 candidate genes. We then shortlisted the top 10 genes for each of the four subtypes using rank sum test (p-value less than 0.001). For two subtypes namely Basal and Her2, three genes had the identical p-value at position 10. So, we incorporated them as well. As a result, we uncovered a set of 41 unique genes by combining the top 10 genes of each subtype. Considering the entire candidate gene set of size 323, we also examined the top ${1/3}^{rd}$ subtype-specific genes(obtained using rank sum test) for each subtype. We discovered two genes shared amongst all subtype-specific gene sets. Thus, after counting these two genes also, we finally arrived at a gene signature of size 43. It was intriguing to discover seven overlapping genes with the PAM50 gene set, namely, 'EXO1', 'ERBB2', 'KRT5', 'CCNE1', 'GRB7', 'CEP55', 'KRT14'. 

\begin{figure}[!htbp]
\centering
	\subfloat[Confusion matrix signifying number of samples correctly identified.]{
	\includegraphics[width=3in,height=4.9cm]{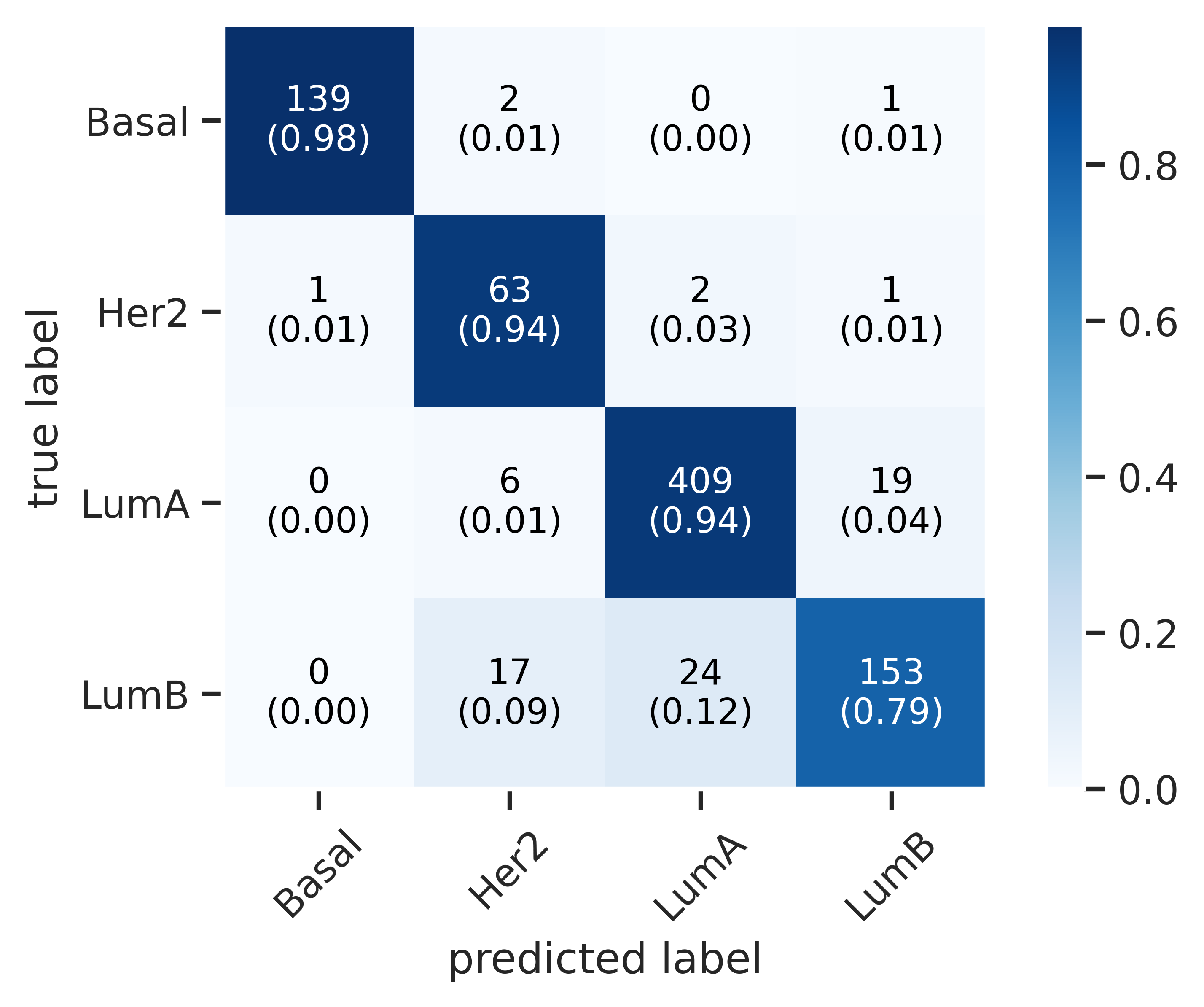}
	\label{fig:06a}
	}
	\hfill
	\subfloat[Heatmap summarizing classification performance metrics for different breast cancer subtypes. Note that despite class-imbalance, the micro-averaged value of each of these metrics is above 0.9]{
	\includegraphics[width=3in,height=5cm]{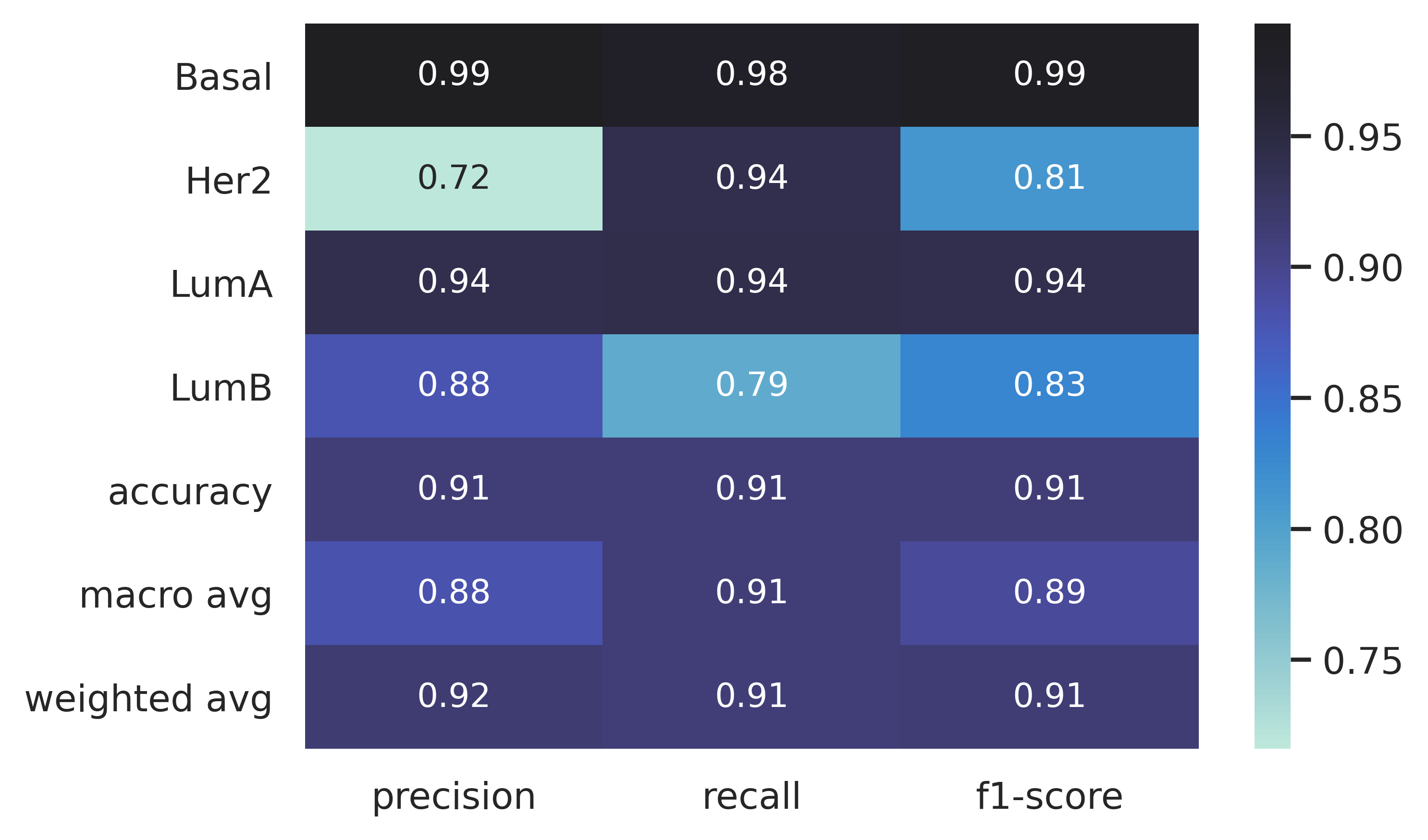}
	\label{fig:06c}
	}
	\\
	\subfloat[The results in box plot depicts the variation in overall accuracy across 10 different folds]{
	\includegraphics[width=3in,height=5.9cm]{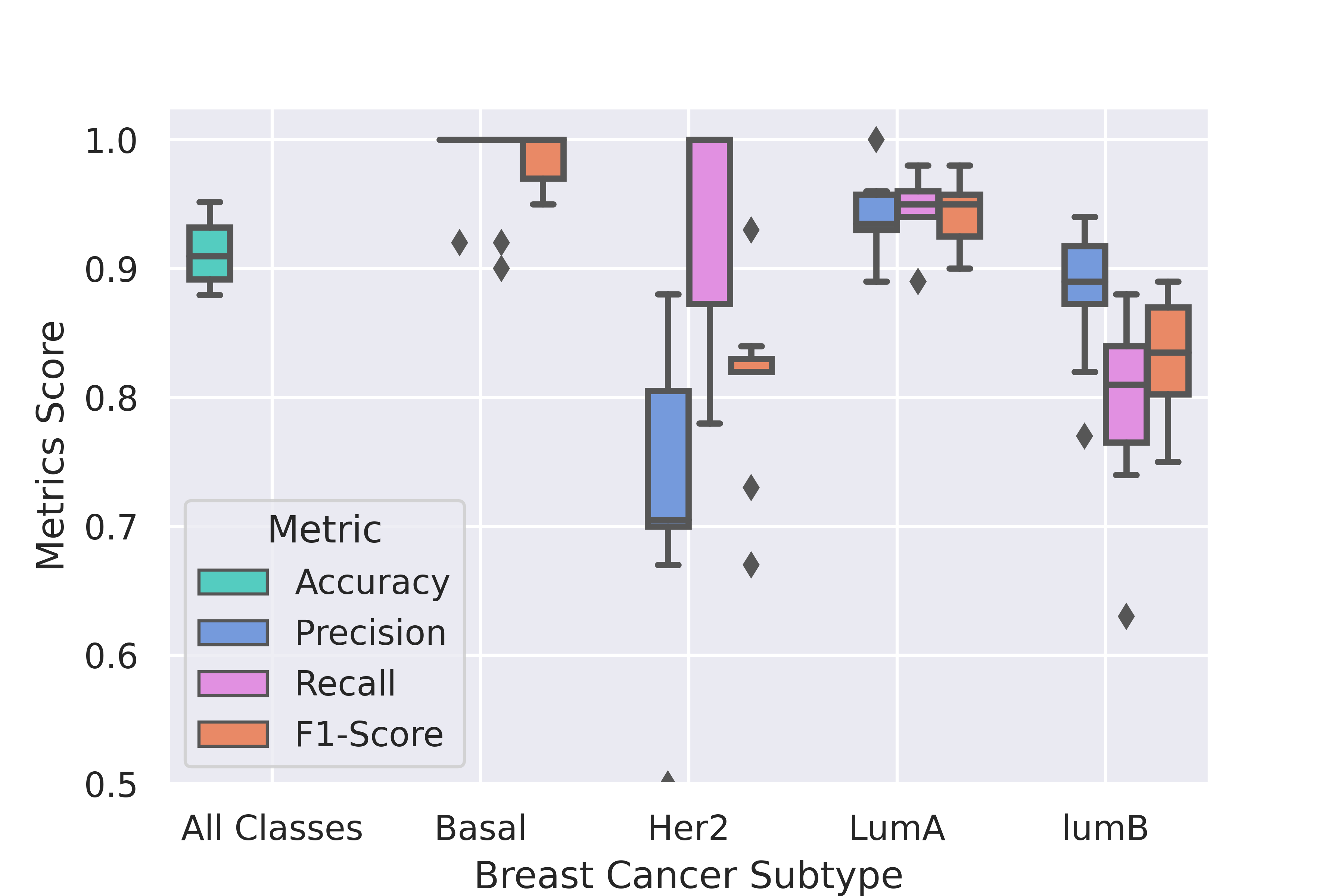}
	\label{fig:06b}
	}
\caption{Classification performance using gene signature of size 43 and 10 fold cross-validation}
\label{fig:06}
\end{figure}

\par We utilized the neural network proposed in \cref{NeuralNet} to assess the capability of identified gene signature set of size 43 to differentiate between four breast cancer subtypes. As shown in the \Cref{fig:06c,fig:06b}, we achieved classification accuracy of 0.91 using 10-fold cross-validation. From the confusion matrix (please see \Cref{fig:06a}), it becomes obvious that using 43 genes, we have been able to correctly classify nearly all samples of Basal subtype. Also, for the basal subtype, we obtained  high scores for precision, recall, and F-score (heatmap in \Cref{fig:06c}). The boxplots (please see \Cref{fig:06b}) show the consistency of these assessment metrics across 10 different runs with the least variation observed for the Basal subtype. As apparent from the results (heatmap in \Cref{fig:06c}), the framework yields a high value ($0.94$) of precision, recall, and F1-measure for Luminal A subtype. On the other hand, these measures have a value approx $0.80$ for the Luminal B subtype. The model scores poorly for the Her2 subtype due to the presence of fewer samples for the subtype in the dataset. Thus, the classification results illustrate that the proposed gene signature set is differentially expressed in four subtypes.

\subsection{Comparison with related work} \label{comparison}

\par More recent literature \cite{zhang2017novel,chen2019identifying,gao2019deepcc} relates the biomarkers to the five subtypes that include another subtype normal like in addition to the four types Basal, Her2, LumA, and LumB. However, for a fair comparison, we compared our work with the one considering four-class classification problem. We summarize the outcome of the proposed model along with work by Zhang et al. \cite{zhang2018lncrna} in terms of accuracy and gene count (see \Cref{Tab:05}).

\par It is evident that for the four subtype classification problem, whereas the proposed model achieves the highest accuracy of 0.91, its competitor \cite{zhang2018lncrna} achieves an accuracy of 0.958. While the proposed model achieves this accuracy using just 43 genes, however, \cite{zhang2018lncrna} requires 106 genes, including 19 PAM50 genes. In their model, when the number of genes is reduced to 36, the classification accuracy falls to 0.885.  

\par We also studied the effect of including the top five PAM50 genes for each of the four subtypes (one vs all) using a rank-sum test. Thus, taking the union of 20 (=$5X4$) genes, with a set of 43 biomarker genes, we obtained a set of 57 unique genes. Using a 57-gene signature, we achieved an accuracy of 0.944, which marks a significant improvement in the performance, although at the cost of a slight increase in the size of the gene set (includes 21 PAM50 genes in total).

\begin{table}[!htpb]
\centering
\caption{Comparison of Classification performance and size of gene signature with state-of-the-art work \label{Tab:05}}
\begin{tabular}{|l|c|c|c|}
\hline
\multicolumn{1}{|c}{\multirow{2}{*}{\textbf{Research Group}}} & \multicolumn{3}{|c|}{\textbf{Result}}             \\
\multicolumn{1}{|c}{}                                  & Omic Data Used & Gene Count & Accuracy \\ \hline
Proposed Model & RNA Sequence Gene Expression                              & 43              & \textbf{0.91}       \\ \hline
\multirow{3}{*}{Zhang et al. \cite{zhang2018lncrna}}                                                  
																	& Coding and non-coding genes & 36 & 0.885  \\
																	& Coding and non-coding genes &  &  \\
																	&  with 19 PAM50 genes & 106 & 0.958 \\ \hline
\end{tabular}
\end{table}

\subsection{Gene Set Analysis of Identified Gene Signature Set of size 43} \label{geneAnalysis}

\par
\begin{figure}
\centering
	\subfloat[Differential expression of subtype specific genes illustrating that the genes selected by proposed algorithm corresponding to different breast cancer types are able to distinguish each subtype from the rest.]{
	\includegraphics[width=3in,height=5cm]{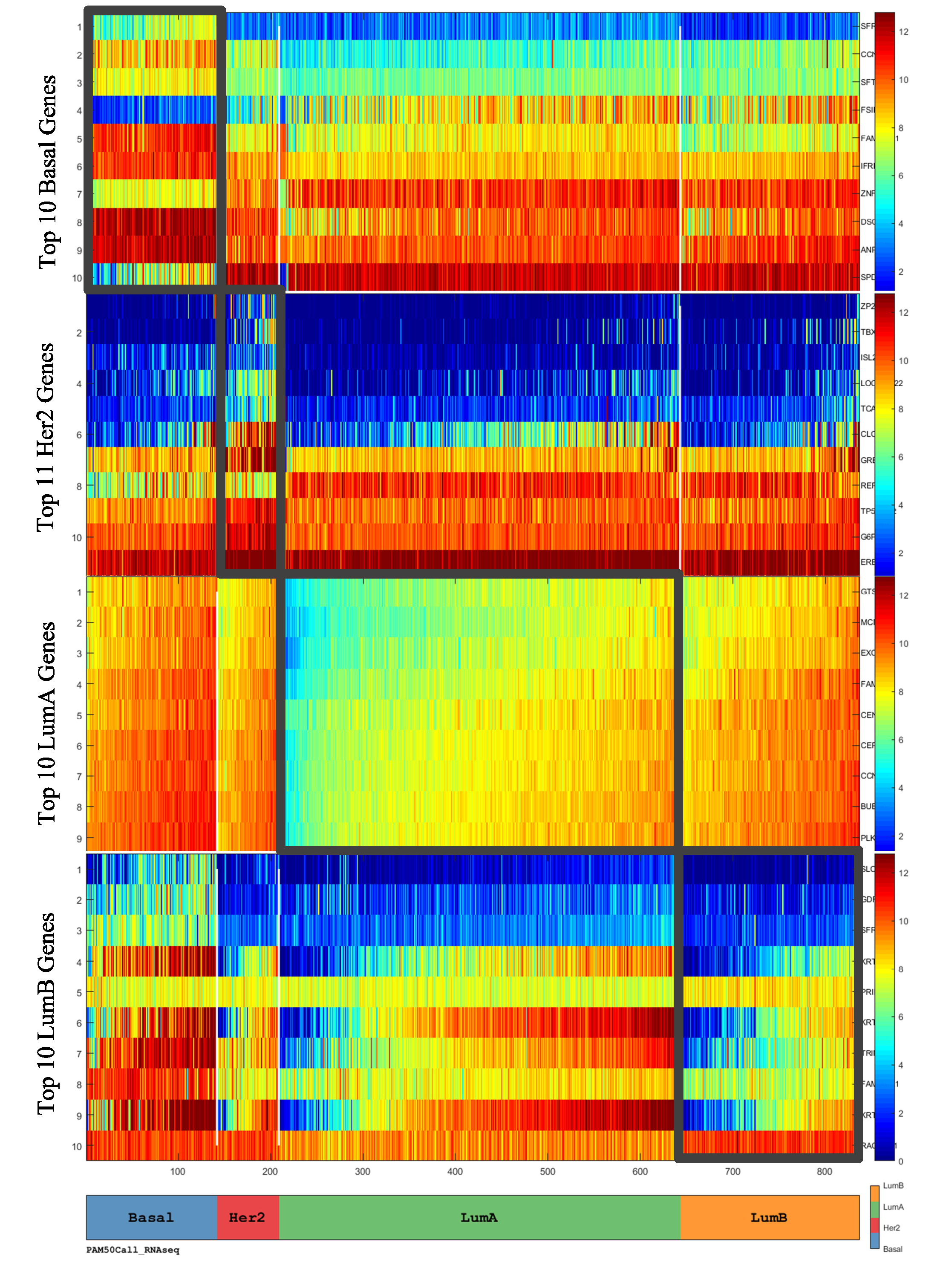}
	\label{fig:07c}}
	\hfill
	\subfloat[T-SNE visualization based on gene signature set of size 43.]{
	\includegraphics[width=3in,height=5cm]{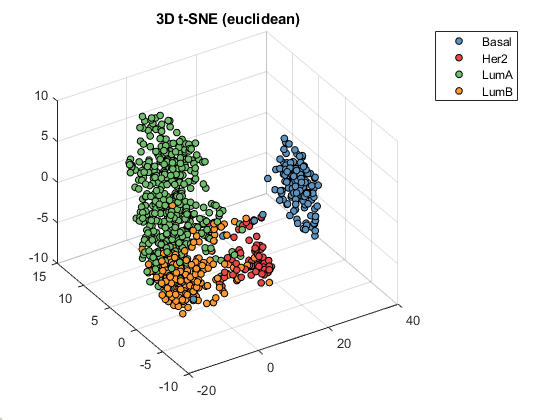}
	\label{fig:07d}}
	\\
	\subfloat[Pearson correlation matrix illustrating high correlation between subtype-specific genes]{
	\includegraphics[width=3in,height=5.5cm]{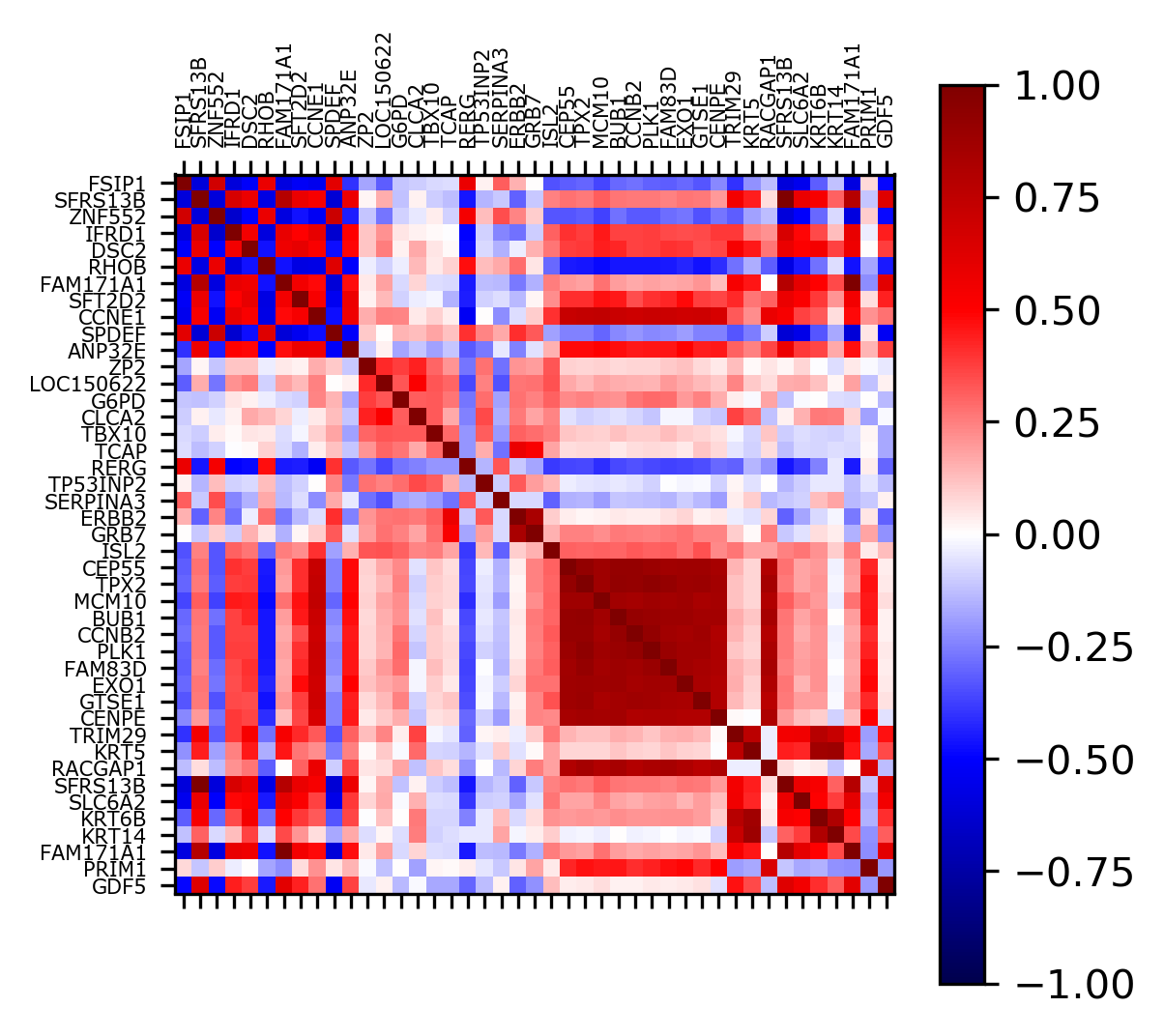}
	\label{fig:07e}}	
\caption{Visual Analysis for the identified 43 biomarker genes}
\label{fig:07}
\end{figure}

 For showing how gene expression level of gene signature of size 43 changes among samples from various breast cancer subtypes, we depicted heatmap in \Cref{fig:07c}. It may be noted that the gene intensity level clearly distinguishes four subtypes. Using the t-distributed Stochastic Neighbor Embedding (t-SNE) approach, we also visualized the gene expression values for identified signature in a reduced three-dimensional space. The segregated clusters highlight the differential expression of genes.  We also investigated the Pearson correlation coefficient between gene groups associated with each of the four subtypes (\Cref{fig:07e}). There exists a high correlation between gene signatures identified for Basal, LumA, and LumB subtypes. Further, it is noteworthy that the genes corresponding to the Luminal A subtype are most closely correlated. 

\begin{figure}
\centering
\subfloat[GO Biological Processes ]{
	\includegraphics[width=4in,height=8.9cm]{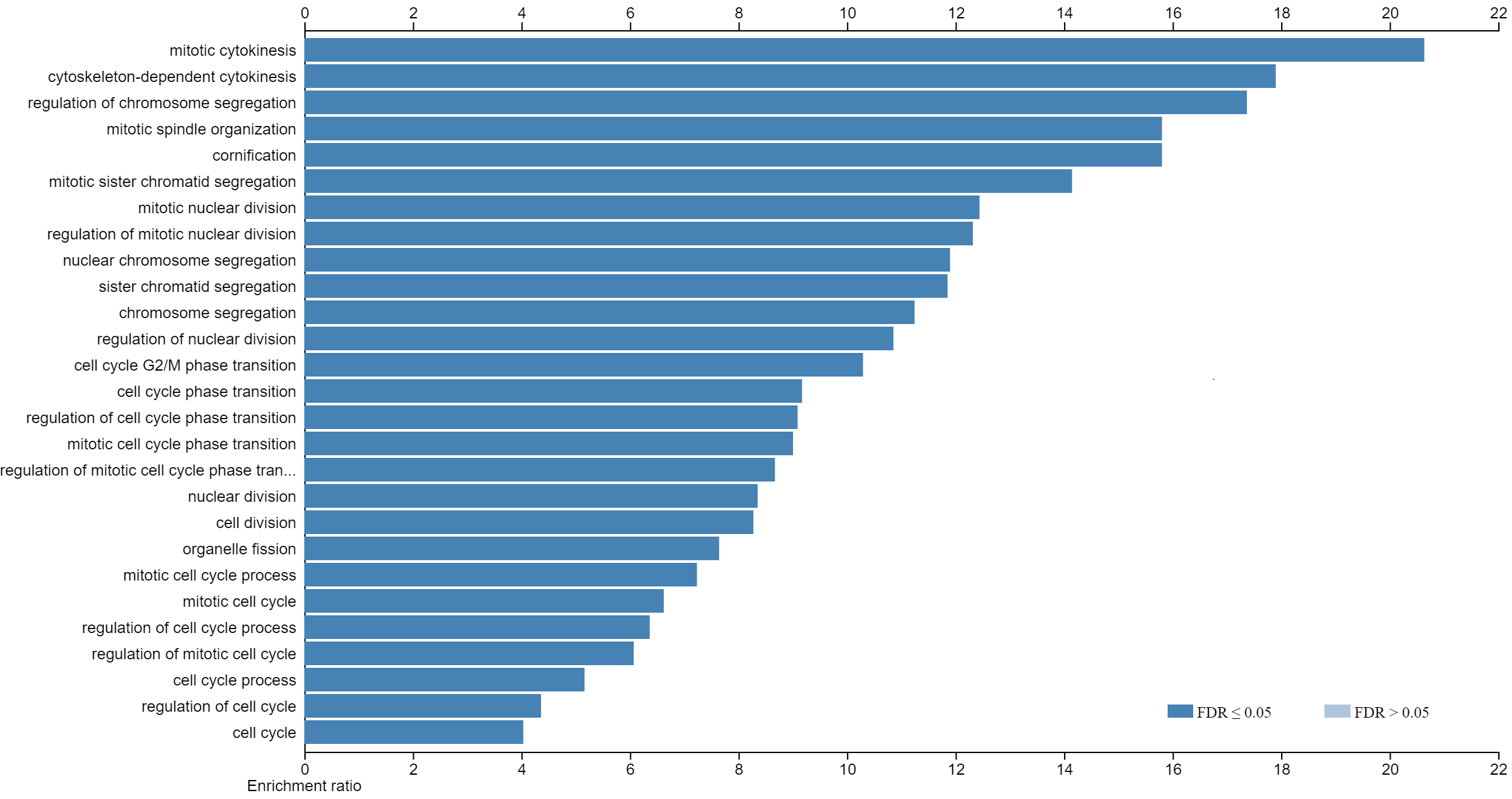}
	\label{fig:09a}}
	\\
	\subfloat[Reactome Pathways]{
	\includegraphics[width=3.2in,height=4.8cm]{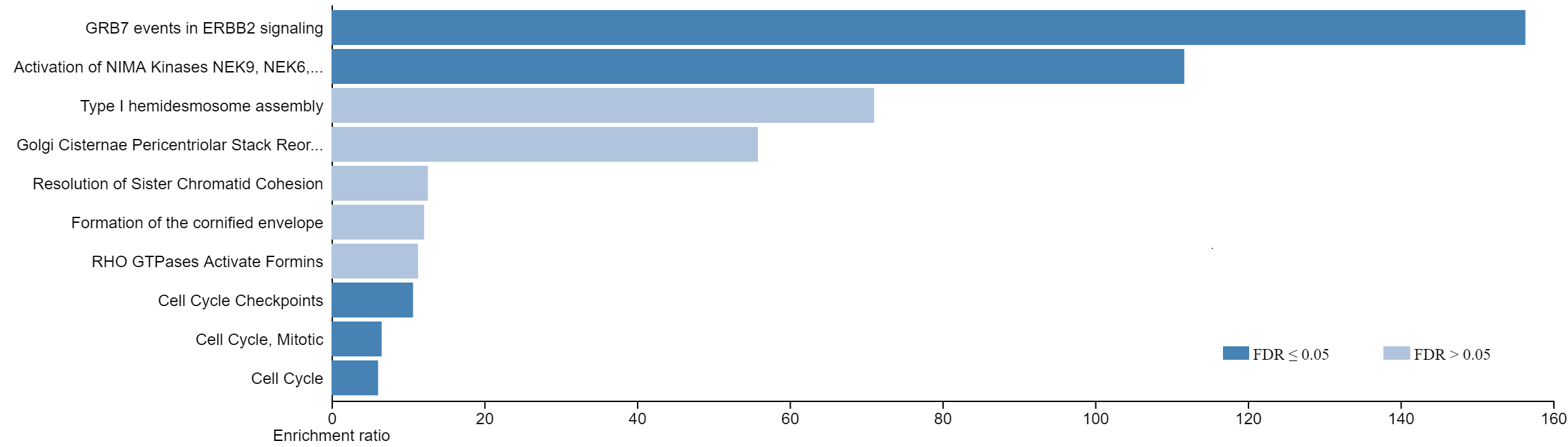}
	\label{fig:09b}}
	\subfloat[Panther Pathways]{
	\includegraphics[width=3.2in,height=4.8cm]{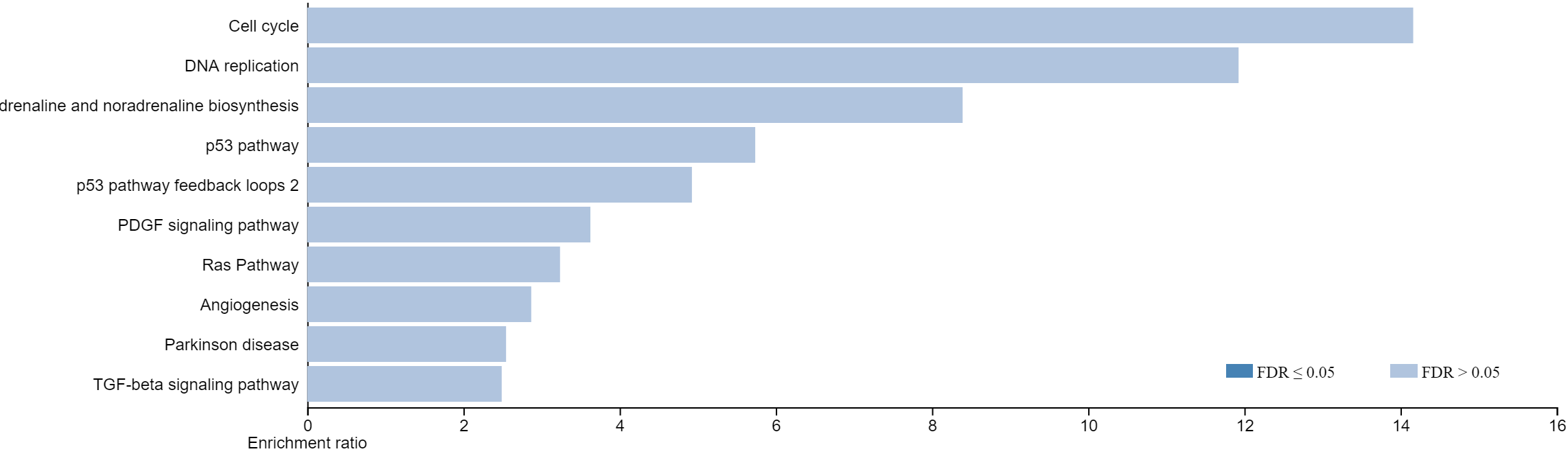}
	\label{fig:09c}}
	\\
	\subfloat[Kegg Pathways]{
	\includegraphics[width=4in,height=3.4cm]{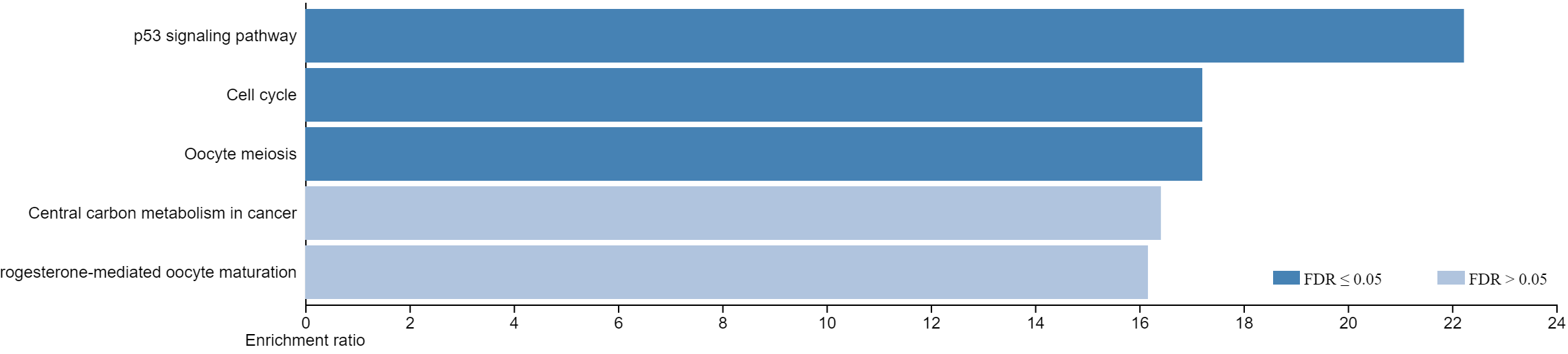}
	\label{fig:09d}}
	
\caption{Analysis of gene signature set of size 43}
\label{fig:09}
\end{figure}

\par Gene ontology corresponding to identified gene signature is shown in \Cref{fig:09a}. To establish the clinical relevance of discovered gene signature, we determine biological processes driven by them. Further, in \Cref{fig:09b,fig:09c,fig:09d}, we have depicted the top 10 Reactome Pathways, top 10 Panther Pathways, and top 5 KEGG Pathways found in over-representation analysis carried out for these genes using WebGestalt tool \cite{liao2019webgestalt}. The pathways being enriched include GRB7 events in ERBB2 signaling, p53 signaling pathway, and Cell Cycle. It may be noted that the discovered gene signature also hit Angiogenesis, Ras pathway, DNA replication, PDGF signaling, and TGF-beta signaling pathways, marked as important pathways for breast cancer in literature.  

\section{Conclusion} \label{sec:conclusion}

\par Breast cancer is a heterogeneous disease having a very high mortality rate that needs utmost attention. Devising appropriate therapy for the treatment of the disease needs identification of subtype-specific gene signature for four breast cancer subtypes, namely Basal, Her2, Luminal A, and Luminal B. In this regard, we proposed an algorithm for gene signature identification for each of these four subtypes of breast cancer. The proposed algorithm examines the trained deep neural network classifier and applies different Explainable AI methods to arrive at a gene signature set of 43 genes. We achieved a mean 10-fold test accuracy of 91.085 percent by feeding these 43 genes into a feed-forward neural network classifier, with weighted average precision, recall, and f1-scores of 0.92, 0.91, and 0.9, respectively. Comparison with similar work highlights competitive performance with respect to accuracy as well as gene signature size.

\par Analysis of the discovered gene set revealed several relevant pathways such as GRB7 events in ERBB2 signaling, p53 signaling pathway, and cell cycle. Using the Pearson correlation matrix, we noted that the subtype-specific genes are correlated within each subtype. In summary, the proposed methodology has enabled us to discover a gene signature set comprising 43 differentially expressed genes across four breast cancer subtypes. Also, the relevance of the discovered gene signature is validated through gene set analysis

\par As part of future work, we aim to study the proposed algorithm for the discovery of gene signature capable of differentiating different types of cancer. Further, we intend to incorporate genome and epigenome data also along with transcriptome data for studying this heterogeneity.

\bibliography{BiomarkerGene_THC}
\end{document}